\documentclass[default,iicol]{sn-jnl}
\usepackage{natbib}
\usepackage{graphicx}
\usepackage{array}     
\usepackage{tabularx}  
\usepackage[utf8]{inputenc}

\theoremstyle{thmstyleone}%
\theoremstyle{thmstyletwo}%

\theoremstyle{thmstylethree}%

\begin{document}

\title[Article Title]{A novel gesture interaction control method for rehabilitation lower extremity exoskeleton}

\author[1]{\fnm{Shuang} \sur{Qiu}}\email{qiushuang@buaa.edu.cn}

\author[1]{\fnm{Zhongcai} \sur{Pei}}\email{peizc@buaa.edu.cn}

\author[2]{\fnm{Chen} \sur{Wang}}\email{wangc@zstu.edu.cn}

\author[2]{\fnm{Jing} \sur{Zhang}}\email{2024210702015@mails.zstu.edu.cn}

\author*[1]{\fnm{Zhiyong} \sur{Tang}}\email{zyt\_76@buaa.edu.cn}

\affil*[1]{\orgdiv{School of Automation Science and Electrical Engineering}, \orgname{Beihang University}, \orgaddress{\city{Beijing}, \postcode{100191}, \country{China}}}

\affil[2]{\orgdiv{School of Information Science and Engineering}, \orgname{Zhejiang Sci-Tech University}, \orgaddress{\city{Hangzhou}, \postcode{310018}, \country{China}}}

\abstract{With the rapid development of Rehabilitation Lower Extremity Robotic Exoskeletons (RLEEX) technology, significant advancements have been made in Human-Robot Interaction (HRI) methods. These include traditional physical HRI methods that are easily recognizable and various bio-electrical signal-based HRI methods that can visualize and predict actions. However, most of these HRI methods are contact-based, facing challenges such as operational complexity, sensitivity to interference, risks associated with implantable devices, and, most importantly, limitations in comfort. These challenges render the interaction less intuitive and natural, which can negatively impact patient motivation for rehabilitation. To address these issues, this paper proposes a novel non-contact gesture interaction control method for RLEEX, based on RGB monocular camera depth estimation. This method integrates three key steps: detecting keypoints, recognizing gestures, and assessing distance, thereby applying gesture information and augmented reality triggering technology to control gait movements of RLEEX. Results indicate that this approach provides a feasible solution to the problems of poor comfort, low reliability, and high latency in HRI for RLEEX platforms. Specifically, it achieves a gesture-controlled exoskeleton motion accuracy of 94.11\% and an average system response time of 0.615 seconds through non-contact HRI. The proposed non-contact HRI method represents a pioneering advancement in control interactions for RLEEX, paving the way for further exploration and development in this field.}

\keywords{Rehabilitation lower extremity robotic exoskeletons, Human-robot interaction, Gesture recognition, Distance estimation,Augmented reality}

\maketitle

\section{Introduction}\label{sec1}

Exoskeleton robots are wearable devices designed to be worn by humans, enabling coordinated movement with the wearer’s limbs to perform tasks. These systems primarily consist of an external framework worn on the body and a powered actuation system. Their distinguishing feature lies in leveraging the powered system to assist the movement of human joints and muscles, thereby enhancing, extending, compensating for, or replacing the functional capabilities of human limbs\cite{bib1}. Applications include performance augmentation, increased endurance, and mobility assistance for individuals with disabilities or the elderly. Unlike traditional rehabilitation devices, which are often limited to use in controlled environments such as hospitals, robotic exoskeleton systems can be used in a broader range of settings. Although extensive research has been conducted on robotic exoskeletons, their control remains a significant challenge\cite{bib2}\cite{bib3}\cite{bib4}. Special attention must be paid to ensuring user-friendly interaction between the exoskeleton and humans in most application scenarios\cite{bib5}\cite{bib6}\cite{bib7}. This is particularly critical for rehabilitation contexts, such as gait retraining for individuals with spinal cord injuries (SCI) or stroke patients\cite{bib8}.  

In rehabilitation training, the patient's active participation in lower limb control is crucial, and they need to gradually complete the training and learning of normal gait through assistance. Identifying the patient's active intention is the key to achieving HRI and rehabilitation movements. Accurate recognition of movement intentions ensures the effectiveness of training, while selecting appropriate signal sources and establishing reliable HRI is essential for enabling the exoskeleton to operate in accordance with the patient's intentions. Currently, physical interaction methods commonly employ mechanical sensors to acquire and process intent signals. Devices such as Inertial Measurement Units (IMU), angular encoders, and foot pressure sensors are used to capture the state of the human-robot system and predict the next movement intention\cite{bib9}.However, these intent signals can only be captured after the subject performs the corresponding movement, resulting in a delay compared to body movements. Furthermore, some motion sensors are bulky and inconvenient to carry, which limits their application in certain scenarios. Some traditional lower-limb rehabilitation exoskeletons control motion modes through operation buttons, joysticks, or wristwatch software. For example, the Atalante system integrates a button controller\cite{bib10}, the Rex system uses a joystick mounted on the waist\cite{bib11}, and the ReWalk Personal 6.0 is operated via a wristwatch\cite{bib12}. While interactions based on motion sensors and mechanical controls are generally stable and easy to identify when switching motion modes, the increasing diversity of motion modes introduces complexity. This not only raises the difficulty of implementing traditional motion sensors, mechanical buttons, joysticks, and wristwatches but also increases the operational burden on users. In lower-limb rehabilitation, accurately recognizing the user's motion intentions under complex movement modes is particularly crucial for patients with paralysis. However, due to weakened muscle strength in the lower limbs, certain contact-based interaction methods—such as motion feedback, force feedback, and bio-electrical signals—are limited. In contrast, the unaffected upper limbs, which have stable motor signals, have become a key focus of research and application in rehabilitation therapy.

To help lower-limb paralysis patients overcome the challenges of using lower-limb electromyographic (EMG) signals, researchers have attempted to utilize EMG signals from the unaffected upper limbs for human motion intention recognition. For example, the MINDWALKER exoskeleton predicts lower-limb movement by leveraging the correlation between arm and leg movements during walking, using upper-limb EMG signals to forecast the motion trajectory of the thighs, calves, and feet\cite{bib13}. Furthermore, EMG signals from upper-limb gestures can provide the necessary motion intentions for multi-mode control of lower-limb rehabilitation exoskeletons, allowing the user to control the exoskeleton through gestures\cite{bib14}.However, as physiological signals, EMG signals are susceptible to interference from factors such as noise\cite{bib15}, sweat\cite{bib16}, and skin impedance\cite{bib17}. There are also inter-individual differences and intra-individual variations\cite{bib18}, which limit the widespread application of EMG. Similarly, other bio-electrical signals, such as electroencephalography (EEG), electrooculography (EOG), and electrocardiography (ECG), face analogous challenges\cite{bib19}, including risks associated with implantation, low signal-to-noise ratios, susceptibility to interference, individual variability, limited signal bandwidth, and difficulty in identifying multiple intentions under complex movement patterns\cite{bib20}. Moreover, the hardware and algorithmic complexities involved in the acquisition and processing of these signals add to the overall cost.

The aforementioned HRI technologies, whether based on IMUs, angle sensors, force sensors, controllers, joysticks, wristwatches, or bio-electrode wearable devices, are all contact-based. These systems often exhibit some limitations in terms of comfort, robustness, and sensitivity, and their ability to accurately reflect human motion intentions is also constrained\cite{bib21}.As a result, non-contact interaction technologies, such as computer vision, have emerged as a promising new direction\cite{bib22}. Computer vision, by mimicking biological vision, enables machines to perceive their surrounding environment. In recent years, it has developed rapidly\cite{bib23}\cite{bib24}, offering a low-cost and easily deployable solution. Similarly, computer vision can recognize users' motion intentions and actions, allowing exoskeleton robots to synchronize with the user's movements and providing a more natural and intuitive interaction experience. Currently, most visual recognition methods are built on deep learning neural networks, which face challenges related to scene recognition and reliability. Factors such as lighting changes and background complexity can significantly affect the quality of images captured by visual sensors and the reliability of intention recognition\cite{bib25}.This makes it more difficult to achieve reliable motion intention capture using visual streaming recognition methods in complex environments.

To address issues such as poor comfort, low reliability, and high latency in the HRI control of rehabilitation lower-limb exoskeletons, this paper proposes a non-contact HRI control method based on computer vision. The method utilizes a three-step fusion algorithm, called Keypoints detection, Gesture classification, and Distance triggering Fusion Algorithm (KGDFA), based on an RGB monocular camera. The algorithm first employs the Mediapipe Hands algorithm\cite{bib26} for hand keypoints detection, optimizing the keypoints connection logic. Then, using the improved keypoints connections, a two-layer recognition convex structure is constructed to identify different hand gestures. For flexion gestures, the algorithm calculates the two-dimensional angles formed by keypoints connections, setting thresholds to classify the gestures and improve recognition accuracy. Finally, the method uses monocular RGB vision to trigger virtual buttons, controlling the exoskeleton to execute corresponding gait actions, thus enhancing interaction safety. In addition, a Finite State Machine (FSM) is designed for the exoskeleton's movements to ensure the safe transition between gait modes. This innovative HRI control method, based on visual detection, recognition, and augmented reality triggering, offers potential applications in fields such as rehabilitation assessment\cite{bib27}\cite{bib28}. 

\subsection{Contribution}\label{subsec2}

The main contributions of our work are summarized as follows:

(1)In light of the challenges posed by the weakened muscle strength of lower-limb paralysis patients and the limited number of recognizable patterns in brainwave signals, this study focuses on gesture-based motion intention recognition. It proposes a novel HRI method, called KGDFA, and establishes a HRI interface for a rehabilitation lower-limb exoskeleton robot platform using an RGB monocular camera.

(2)The developed HRI control system, KGDFA, relies on hand keypoints detection, a two-layer recognition convex framework, and constraints on the two-dimensional joint angles of fingers to interpret gestures. This approach addresses the challenge of recognizing multiple intentions under complex motion patterns at a relatively low cost.

(3)KGDFA employs distance estimation using a low-cost and widely applicable monocular RGB camera to enable virtual button triggering, thereby avoiding the complex computations and potential failures associated with using depth cameras to measure hand distance. On one hand, this approach simplifies the system's signal processing complexity and improves real-time performance. On the other hand, it effectively addresses the low reliability often associated with visual recognition-based interactions, thereby ensuring the safety of HRI.

(4)The gesture-controlled rehabilitation lower-limb exoskeleton robot achieved a motion execution accuracy of 94.11\%, demonstrating its effectiveness and reliability, making it more feasible for engineering applications. The gesture control system's average response time of approximately 0.615 seconds indicates its capability for real-time control and execution.

(5)FSM was designed for the actions of the rehabilitation lower-limb exoskeleton, further enhancing the safe transition of gait motion modes under gesture-based interaction.

\subsection{Organization}\label{subsec2}

The remainder of this paper is organized as follows. Section 2 reviews related work on gesture recognition and interaction control. The proposed KGDFA and the FSM transition strategy for exoskeleton gait movements are detailed in Section 3. Experimental evaluation results are presented in Section 4. Finally, conclusions and directions for future work are discussed in Section 5.

\section{Related work}\label{sec2}

Gesture recognition is a popular research topic in the HRI of RLEEX, playing a crucial role in the interaction between humans and exoskeletons\cite{bib29}. Two of the most widely used gesture recognition techniques are surface electromyography (sEMG)-based gesture recognition and vision-based gesture recognition. sEMG-based gesture recognition works by integrating and processing different sEMG signal data\cite{bib30}; however, the generalizability of sEMG signal pattern recognition systems is limited. This limitation is particularly evident in the system's susceptibility to errors caused by factors such as electrode displacement, muscle fatigue, and variations in muscle contraction strength. Additionally, there are inter-individual differences in EMG signals, as well as intra-individual variability over time, which restricts the widespread application of EMG signals. In contrast, vision-based gesture recognition relies on visual data captured by sensors to identify gestures. This method leverages deep learning neural networks to classify gestures by learning from large datasets of gesture images across diverse populations, offering excellent generalizability across different user groups. Moreover, it is a non-contact approach, completely avoiding the discomfort that may arise from sensor-to-skin contact, which makes the HRI process more natural and comfortable. Xu et al.\cite{bib31} proposed an accurate hand detection method based on a hybrid detection or reconstruction convolutional neural network (CNN) framework, achieving reliable detection of multiple hands in cluttered backgrounds. Ibanez et al.\cite{bib32} introduced the EasyGR gesture recognition tool, which is based on machine learning algorithms and allows developers to customize and recognize gestures without needing deep knowledge of machine learning algorithms, thus reducing the development workload. Mujahid et al.\cite{bib33} presented a lightweight gesture recognition model based on YOLOv3 and the DarkNet-53 CNN\cite{bib34}, which requires no additional preprocessing and can recognize gestures even in complex environments with low resolution.

With the widespread adoption of gesture recognition technology, academia is exploring its applications in fields such as medicine, construction, and robotics to enhance labor productivity\cite{bib35}. Simple gestures allow users to interact with devices in the air without physical contact. In control system interactions, gestures are considered safer and easier than button-based controls\cite{bib36}.Rudd et al.\cite{bib37} combined gesture control technology with robotic motion control techniques to achieve accurate gesture recognition, enabling control over the robot's position and speed. Additionally, the system integrates obstacle avoidance functionality, ensuring the robot's safety and stability. Neto et al.\cite{bib38} proposed a gesture-based HRI system that applies gesture recognition technology in manufacturing, assisting operators in assembly tasks. Mazhar et al.\cite{bib39}\cite{bib40} developed a comprehensive skeletal extraction and gesture recognition process, applying it to real-time physical interaction between humans and robots. Through experimental interaction, the performance of the proposed framework was validated. This framework combines depth information extracted from Kinect with skeletal extraction from OpenPose\cite{bib41} to determine the spatial position of the hand. Using a convolutional neural network, the system recognizes gesture information, allowing the robot to interpret human commands and perform tasks such as tool exchange between humans and robots. Shi et al.\cite{bib42} proposed a real-time gesture-controlled robotic 3D printing HRI method. This method uses YOLOv5\cite{bib43} and Mediapipe algorithms to recognize gestures and translate gesture information into real-time robotic operations. Experimental results demonstrated the effectiveness and reliability of gesture control. Regarding visual gesture recognition for exoskeleton interaction control, Baulig et al.\cite{bib44} evaluated the performance of a robot-controlled, hand-based exoskeleton's monocular hand posture estimation in simulation. However, their study did not conduct experiments on real exoskeleton prototypes.

Currently, gesture control technology faces challenges such as low precision, complex operation, and prolonged recognition time. The accuracy of gesture recognition directly impacts the precision of gesture control. This study aims to improve existing gesture recognition algorithms to enhance their accuracy and real-time performance; And introduces augmented reality triggering based on depth estimation of monocular RGB camera, which sets triggering conditions with depth threshold to improve the reliability of interaction; Thus ensuring the quality of gesture-controlled RLEEX.

\section{Methodology}\label{sec3}

The following statement is made for this paper: (1) All methods were carried out in accordance with relevant guidelines and regulations. (2) All experimental protocols in our paper were approved by the School of Automation Science and Electrical Engineering of Beihang University. (3) Informed consent was obtained from all subjects.

This study focuses on controlling exoskeleton robots to perform specific movements through gesture control. The objective is to obtain digitized gesture results via gesture recognition software and convert them into appropriate control commands. First, we designed and implemented hand keypoints detection based on MediaPipe Hands to distinguish the interactive hand from the background. On this basis, we improved the keypoints connection logic to better support gesture recognition and introduced a virtual keypoint to simplify the overall palm computation. Second, we developed a gesture recognition rule-based algorithm leveraging the improved keypoints connection logic, enabling the classification of 12 distinct gestures, each corresponding to a specific gait pattern of the exoskeleton. Furthermore, we designed a virtual button system based on monocular depth estimation, which requires both a specific gesture and a press-and-release action of the virtual button to control the exoskeleton's execution of the corresponding gait task. This approach ensures reliable HRI while supporting multitasking scenarios. Finally, a FSM gait transition strategy for the exoskeleton gait action is designed to achieve the goal of augmented reality interactive control of the exoskeleton motion mode.

\begin{figure*}[t!]
    \centering
    \includegraphics[width=0.7\textwidth]{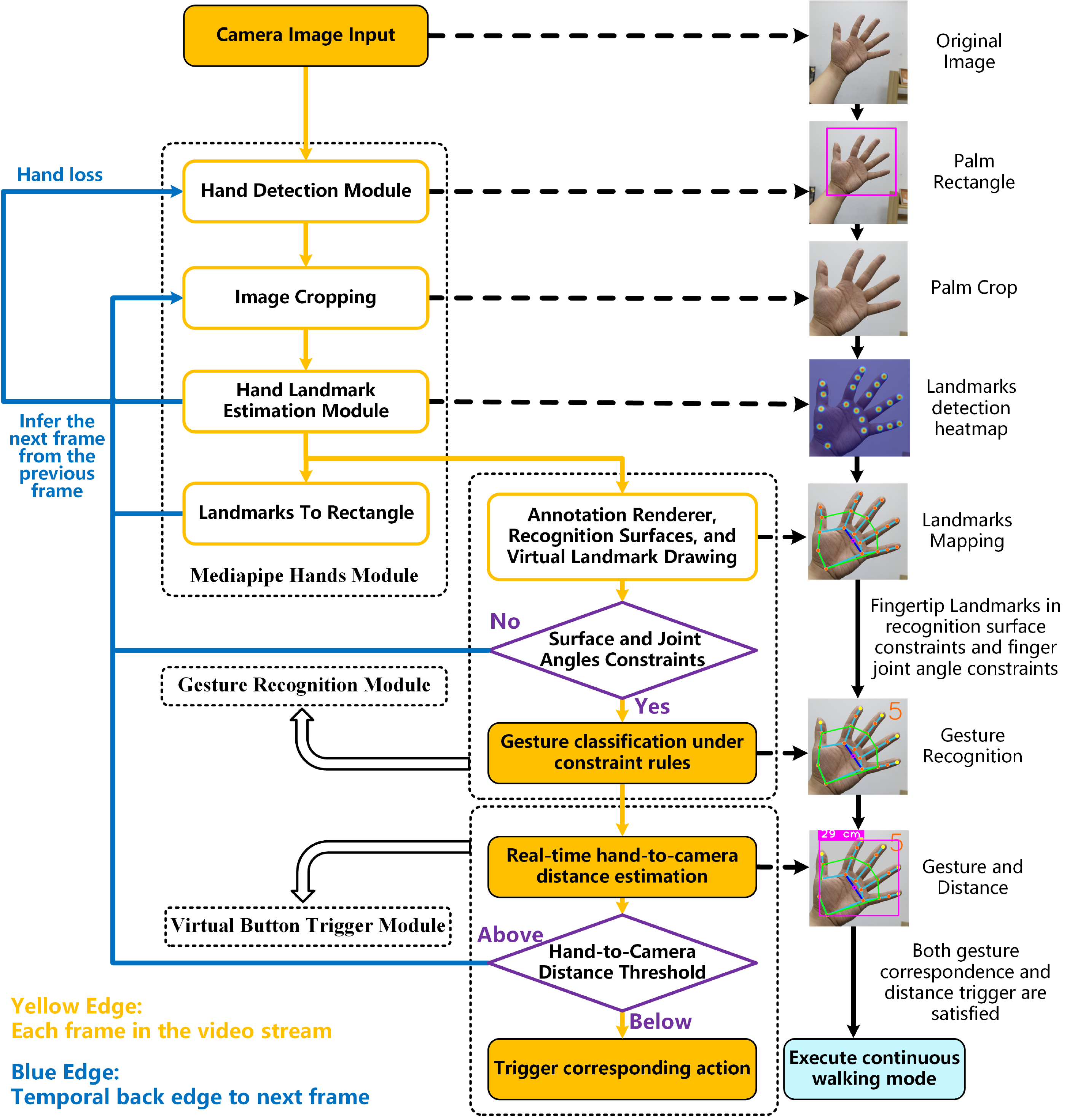}  
    \caption{Flowchart of keypoints detection gesture classification and distance triggering three-step fusion algorithm}
    \label{fig1}
\end{figure*}

The workflow of the proposed Keypoints detection Gesture classification and Distance triggering three-Step Fusion Algorithm (KGDFA) is illustrated in Fig.\ref{fig1}. The algorithm consists of several preprocessing and analysis steps to ensure high-quality results. First, the raw images are preprocessed through denoising and sharpening to reduce data redundancy and complexity while enhancing the detectability of useful information within the images. Next, a palm detection module is used to identify the gesture region, cropping out redundant parts to obtain a refined gesture image. Subsequently, a hand keypoints estimation module is employed to generate joint heatmaps, from which the positions of joint nodes are determined to extract keypoints data and identify the hand skeleton. Then, a two-layer recognition convex surface, specific key points of the five fingertips, finger flexion degree and two-dimensional joint node constraints of pixel distance ratio are introduced to realize gesture recognition. This workflow automatically crops images and achieves high-precision, high-reliability gesture information recognition. Finally, the real-time perpendicular distance between the palm and the camera is calculated to determine whether to trigger a virtual button, thereby executing the exoskeleton gait task corresponding to the recognized gesture.

\subsection{Mapping relationship between gestures and exoskeleton actions}\label{subsec2}

In the context of gesture recognition for controlling exoskeletons, two fundamental prerequisites for effective gesture detection are: (1) the gesture must be recognizable by the camera, and (2) the operator should be able to easily learn, perform, and recall the gesture. Ensuring that gestures are simple to recognize, execute, and replicate is essential for reliable operation. To accommodate the diverse gait modes of the exoskeleton, we customized 12 specific gestures. These include ten numerical gestures (0, 1, 2, 3, 4, 5, 6, 7, 8, 9) and two symbolic gestures (“love” and “rock”). The distribution of these specific gesture categories and their corresponding gait modes is shown in Tab.\ref{Tab1}. As illustrated, these gestures are straightforward, systematic, and easy to recognize.

To enhance the model's ability to learn and understand gesture information, we drew inspiration from the open-source datasets MSRA Hand Tracking database\cite{bib45} and MSRA Hand Gesture database\cite{bib46}, which include diverse gesture images captured from various angles and orientations. Each hand was annotated with 21 keypoints, representing landmarks (i.e., the coordinates of points within the images). The annotated dataset was then randomly divided into 80\% training data and 20\% testing data. After training and evaluation, the final model was obtained.

\subsection{Gesture recognition algorithm based on hand keypoints detection}\label{subsec2}

This study leverages the MediaPipe Hands algorithm\cite{bib18} for hand keypoints detection and tracking. The algorithm consists of two primary modules: a palm detection module and a hand keypoints estimation module. Due to the flexibility of fingers, direct recognition is challenging. Therefore, the workflow is divided into two steps. First, the algorithm uses the relative rigidity of the palm to establish a palm detector, identifying the palm as a preliminary step. Second, after detecting the palm, the specific finger keypoints are identified, enabling more accurate gesture detection. Once the palm detection module detects a hand, subsequent detection frames are handled by the hand keypoints estimation module. This module estimates the palm position by calculating the hand keypoints coordinates from the previous frame and feeds the position to the next frame’s hand keypoints estimation module, eliminating the need to repeatedly invoke the palm detection module for every frame. Additionally, the hand keypoints estimation module outputs a scalar to determine the confidence level of the presence of a hand. If the confidence level of the estimated palm drops below a certain threshold, the palm detection module is reactivated for the next frame. In other words, the output of the hand keypoints estimation module controls when the palm detection module is triggered. This behavior is facilitated through synchronized parallel components, such as a real-time rate limiter, palm image cropping module, keypoints-based palm position estimation module, and keypoints rendering module. These components collectively ensure high performance and optimal throughput for the machine learning pipeline. This mechanism significantly improves the model’s processing speed, making the system more efficient and effective for real-time applications that demand rapid and accurate processing.

\begin{figure*}[t!]
    \centering
    \includegraphics[width=0.7\textwidth]{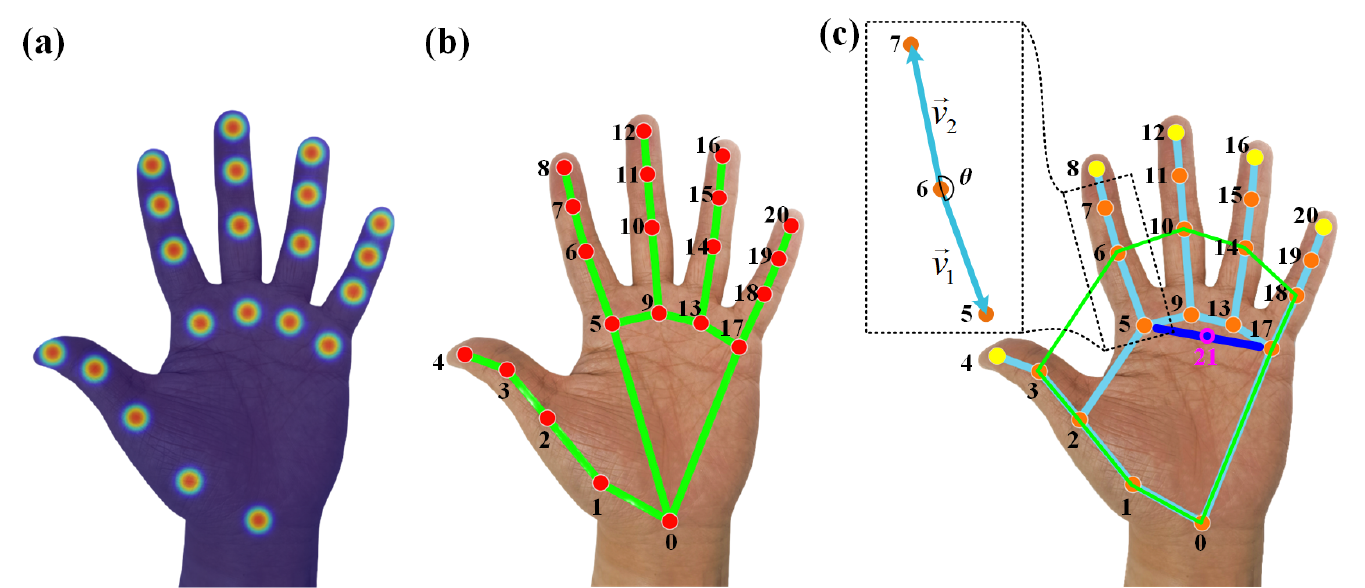}  
    \caption{Identification diagram:(a) Heatmap;(b) MediaPipe skeleton model;(c) Identify convex surfaces, virtual key points, and knuckle angles}
    \label{fig2}
\end{figure*}

The hand keypoints estimation module predicts and generates heatmaps for 21 hand keypoints, from which the coordinates of these keypoints are extracted. By numbering and connecting these 21 keypoints in a specific order, a skeletal model of the hand is formed, as illustrated in Fig.\ref{fig2}(a) and Fig.\ref{fig2}(b).To construct the palm plane, MediaPipe connects keypoints 0, 5, 9, 13, and 17 sequentially. However, the resulting area is relatively small, which can lead to misrecognition of various gestures, adversely affecting gesture interpretation and subsequent control tasks.

\begin{table*}[h!]
\centering
\small  
\caption{Correspondence between different gestures and gait movement patterns}  
  \label{Tab1}
  \renewcommand{\arraystretch}{1.5}  
  \begin{tabularx}{\textwidth}{>{\centering\arraybackslash}m{4cm}|>{\raggedright\arraybackslash}m{7cm}|>{\centering\arraybackslash}m{4cm}}  
  \hline
  \textbf{Gesture Name} & \textbf{Meaning of gestures} & \textbf{Corresponding gesture pictures} \\
  \hline
  Gesture 0 & Initialization:Power failure of joint motor or Power on to enable standing & \includegraphics[width=1cm]{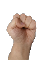} \\ 
  \hline
  Gesture 1 & Step-by-step walking mode:Initial right leg step(Starting gait)  & \includegraphics[width=1cm]{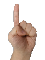} \\
  \hline
  Gesture 2 & Step-by-step walking mode:Walking with left leg(Forward gait) & \includegraphics[width=1cm]{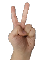} \\
  \hline
  Gesture 3 & Step-by-step walking mode:Walking with right leg(Forward gait) & \includegraphics[width=1cm]{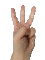} \\
  \hline
  Gesture 4 & Step-by-step walking mode:End the walk by pulling your legs back(End gait) & \includegraphics[width=1cm]{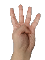} \\
  \hline
  Gesture 5 & Continuous walking mode:Initial left leg step(Starting gait)
The left and right legs walk in a cycle and wait for the command to return to the standing position & \includegraphics[width=1cm]{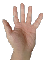} \\
  \hline
  Gesture 6 & Return to standing position in continuous walking mode & \includegraphics[width=1cm]{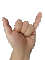} \\
  \hline
  Gesture 7 & Exoskeleton performs stair climbing gait & \includegraphics[width=1cm]{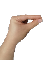} \\
  \hline
  Gesture 8 & Exoskeleton performs stair-descent gait & \includegraphics[width=1cm]{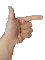} \\
  \hline
  Gesture 9 & Exoskeleton performs obstacle-crossing gait & \includegraphics[width=1cm]{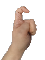} \\
  \hline
  Gesture love & Exoskeleton performs sitting action & \includegraphics[width=1cm]{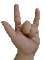} \\
  \hline
  Gesture rock & Exoskeleton performs standing action & \includegraphics[width=1cm]{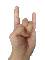} \\
  \hline
\end{tabularx}
\end{table*}

After designing the gesture-task state mapping and implementing hand keypoints detection and tracking, our goal is to accurately recognize gestures while distinguishing the operator from the background. To achieve this and address the limitation of the small hand-palm region, we improved the keypoints connection logic based on MediaPipe Hands. Specifically, we connected keypoint 2 to keypoint 5 and extended the approximate triangular palm region formed by keypoints 0, 5, 9, 13, and 17 into a quadrilateral region by incorporating keypoints 0, 1, 2, 5, 9, 13, and 17. This new region was designated as the inner recognition convex hull. Subsequently, we constructed a larger polygonal area using keypoints 0, 1, 2, 3, 6, 10, 14, 18, and 17, which was designated as the outer recognition convex hull. The construction of these inner and outer convex hulls facilitates gesture recognition by providing a clear reference. By determining which of the five fingertip keypoints (4, 8, 12, 16, 20) fall within the inner or outer convex hulls, we designed a set of gesture actions, as specified in Tab.\ref{Tab2}. Additionally, for gestures involving finger flexion, we calculated the degree of finger bending by analyzing the two-dimensional joint angles formed by connecting the keypoints along each finger. Thresholds for the angles between the upper and lower finger joints were set to classify gestures dynamically, thereby further enhancing recognition accuracy. The improved design is illustrated in Fig.\ref{fig2}(c).

To ensure the accuracy of finger flexion estimation, we set threshold values $\theta_0$ for the angles between the upper and lower joints of the fingers. The vector angles $\theta$ between skeletal points of the gesture are calculated using Eq.\ref{equ1}:
\begin{equation}
\theta=\arccos \left(\frac{v_{1} \cdot v_{2}}{\left\|v_{1}\right\| \cdot\left\|v_{2}\right\|}\right)
\label{equ1}
\end{equation}

The vectors $v_1$ and $v_2$ are illustrated in Fig.\ref{fig2}(c). Here, the fingers are defined to be bent for $\theta$$\le$$\theta_0$ and open for $\theta$$>$$\theta_0$. During the process of determining whether a finger is bent, it was observed that the thumb, forefinger, and pinky have distinct threshold values compared to the other three fingers. Based on extensive experimentation, the threshold angle for thumb flexion was finalized at 53°. For the forefinger, the flexion angle threshold distinguishing Gesture 0 from Gesture 6 was set at 65°, while the thresholds for Gesture 5, Gesture 9, and other Gestures were determined to be 170°, 120°, and 160°, respectively. The threshold for pinky flexion was set at 49°, and the thresholds for the other fingers were set at 65°. 

In addition, Gestures 6, 7, and 8 require the computation of pixel distances between the fingertip keypoints 4, 8, 12, and 20 as a constraint. Assuming the pixel coordinates of two distinct fingertip keypoints in the image are $P^{p i}=\left[u_{1}, v_{1}\right]^{\mathrm{T}}$ and $P^{p j}=\left[u_{2}, v_{2}\right]^{\mathrm{T}}$, the pixel distance between them is calculated as follows:
\begin{equation}
P^{p i j}=\sqrt{\left(u_{2}-u_{1}\right)^{2}+\left(v_{2}-v_{1}\right)^{2}}
\label{equ2}
\end{equation}

Through extensive experimentation, the pixel distance constraints for specific gestures were determined as follows: for Gesture 8, the constraint is $P^{p 48}>100$; for Gesture 6, the constraint is $P^{p 420}>100$. For Gesture 7, a proportional constraint was established based on the distances between keypoints 4 and 12, as well as 8 and 12:
\begin{equation}
\frac{P^{p 412}}{P^{p 812}}<2
\label{equ3}
\end{equation}

\begin{table*}[h!]
\centering
\small  
\caption{Finger key point states and flexion thresholds corresponding to different gestures}  
  \label{Tab2}
  \renewcommand{\arraystretch}{1.6}  
  \begin{tabularx}{\textwidth}{>{\centering\arraybackslash}m{1.6cm}|>{\raggedright\arraybackslash}m{3.5cm}|>{\centering\arraybackslash}m{1.6cm}|>{\centering\arraybackslash}m{1.6cm}|>{\centering\arraybackslash}m{1.6cm}|>{\centering\arraybackslash}m{1.6cm}|>{\centering\arraybackslash}m{1.6cm}}  
  \hline
  \textbf{Gesture Name} & \textbf{Key point status} & \textbf{4 Thumb} & \textbf{8 Forefinger} & \textbf{12 Middlefinger} & \textbf{16
Ringfinger} & \textbf{20 Pinky} \\
  \hline
  Gesture 0 & 4, 8, 12, 16, and 20 are all within the inner convex hull & $<53^{\circ}$ & $<65^{\circ}$ & $<65^{\circ}$ & $<65^{\circ}$ & $<49^{\circ}$\\ 
  \hline
  Gesture 1 & Only 8 is outside the outer convex hull  & $<53^{\circ}$ & $>160^{\circ}$ & $<65^{\circ}$ & $<65^{\circ}$ & $<49^{\circ}$ \\
  \hline
  Gesture 2 & Only 8 and 12 are outside the outer convex hull & $<53^{\circ}$ & $>160^{\circ}$ & $>65^{\circ}$ & $<65^{\circ}$ & $<49^{\circ}$ \\
  \hline
  Gesture 3 & 8, 12, and 16 are outside the outer convex hull & $<53^{\circ}$ & $>160^{\circ}$ & $>65^{\circ}$ & $>65^{\circ}$ & $<49^{\circ}$ \\
  \hline
  Gesture 4 & 8, 12, 16, and 20 are outside the outer convex hull & $<53^{\circ}$ & $>160^{\circ}$ & $>65^{\circ}$ & $>65^{\circ}$ & $>49^{\circ}$ \\
  \hline
  Gesture 5 & 4, 8, 12, 16, and 20 are all outside the outer convex hull & $>53^{\circ}$ & $>170^{\circ}$ & $>65^{\circ}$ & $>65^{\circ}$ & $>49^{\circ}$ \\
  \hline
  Gesture 6 & 4 and 20 are outside the outer convex hull and their pixel distance from each other is greater than the set threshold & $>53^{\circ}$ & $<65^{\circ}$ & $<65^{\circ}$ & $<65^{\circ}$ & $>49^{\circ}$\\
  \hline
  Gesture 7 & 4, 8, and 12 are outside the outer convex hull, and their pixel distance ratio from each other is smaller than the setthreshold & $>53^{\circ}$ & $>160^{\circ}$ & $>65^{\circ}$ & $<65^{\circ}$ & $<49^{\circ}$ \\
  \hline
  Gesture 8 & 4 and 8 are outside the outer convex hull, and their pixel distance from each other is greater than the set threshold & $>53^{\circ}$ & $>160^{\circ}$ & $<65^{\circ}$ & $<65^{\circ}$ & $<49^{\circ}$ \\
  \hline
  Gesture 9 & Only 8 is outside the outer convex hull & $<53^{\circ}$ & $<120^{\circ}$ & $<65^{\circ}$ & $<65^{\circ}$ & $<49^{\circ}$ \\
  \hline
  Gesture love & 4, 8, and 20 are outside the outer convex hull & $>53^{\circ}$ & $>160^{\circ}$ & $<65^{\circ}$ & $<65^{\circ}$ & $>49^{\circ}$ \\
  \hline
  Gesture rock & Only 8 and 20 are outside the outer recognition convex hull & $<53^{\circ}$ & $>160^{\circ}$ & $<65^{\circ}$ & $<65^{\circ}$ & $>49^{\circ}$ \\
  \hline
\end{tabularx}
\end{table*}

Tab.\ref{Tab2} presents the thresholds for gesture flexion levels, defined based on the two-layer recognition convex hull and two-dimensional joint angle constraints for finger keypoints. These thresholds enable the gesture recognition software to accurately detect which layer of the recognition convex hull each fingertip keypoint is located in. Additionally, it identifies the corresponding gesture by analyzing the degree of finger flexion and the pixel distance ratios between fingertip keypoints.

After implementing the gesture recognition algorithm, our next objective shifted toward ensuring reliable gesture operations while maintaining smooth interactions and a wide range of interaction capabilities. To achieve this, we enhanced the MediaPipe Hands framework by introducing a virtual keypoint, referred to as keypoint 21. This keypoint 21 is defined as the midpoint of the line segment connecting keypoints 5 and 17, representing the center of the palm’s width (palm width), as illustrated in Fig.\ref{fig2}(c). We designate this virtual keypoint 21 as the anchor point of the palm, which simplifies the computation of all key points on the hand. Furthermore, it serves as the basis for estimating the overall depth of the hand in subsequent steps.

\subsection{Depth estimation-based method for augmented reality interaction triggering}\label{subsec2}

We identify 22 key points of the hand on the pixel plane and determine their maximum and minimum coordinates. Using these extrema, we construct a rectangular region representing the detected hand area, referred to as the bounding box. When the distance between the hand and the camera remains below a predefined threshold, we check whether the virtual button we set is located within the bounding box. If it is, the virtual button is considered pressed (its color changes to green). Conversely, if the hand-camera distance exceeds the threshold or the hand’s bounding box moves away from the virtual button, the button is released (its color changes to red). This mechanism enables the control of virtual button press-and-release functionality in augmented reality interactions. This approach enhances operational certainty and reliability when triggering specific task commands, serving as a safety mechanism. Compared to traditional mechanical button interactions, both methods ensure stable and reliable signals during mode switching. However, with the increasing diversity of motion modes, mechanical buttons require additional physical buttons to accommodate new needs. This leads to an increase in the number of physical interfaces, adding bulk, weight, and complexity to the system while raising the difficulty of user operation. In contrast, the depth estimation-based augmented reality interaction triggering method allows the integration of gestures to correspond to diverse motion modes and task actions. This approach not only ensures reliable HRI but also meets the demands of multitasking scenarios, offering a more streamlined and versatile solution.

The distance between an object and the camera can be obtained using either a binocular camera or a monocular camera. A binocular camera estimates depth by combining baseline, focal length, and disparity through triangulation, enabling it to acquire depth information for any pixel in a video frame with high precision and low error. In contrast, a monocular camera determines the perpendicular distance from the object to the camera based on the pinhole imaging principle. Although it cannot provide depth information for specific pixels and its accuracy is more affected by changes in object posture, it offers advantages in cost, accessibility, applicability, real-time performance, and stability compared to depth cameras. In close-range hand interaction scenarios, the hand's pixel size on the imaging plane allows the use of monocular estimation to determine the distance of the hand from the camera. 

The similar triangles method is a commonly used approach for distance measurement with monocular cameras. Suppose we have a target object of height \textit{H}, placed at a distance \textit{D} from the camera. After capturing the object and measuring its pixel height \textit{h} in the image, and given the camera's focal length \textit{f}, the perpendicular distance from the monocular camera to the object can be determined using the principle of similar triangles:
\begin{equation}
D=\frac{f \cdot H}{h}
\label{equ4}
\end{equation}

For distance-triggered interaction using a monocular RGB camera, we utilize the distance between hand key points 5 and 17 as the real-world anchor size for monocular depth estimation. This distance approximately corresponds to the palm width minus the width of one finger (about one-quarter of the palm width). By inputting the user's gender and height, the corresponding palm width can be obtained from the Chinese Adult Human Body Dimensions dataset\cite{bib47}, allowing the calculation of the real-world anchor size \textit{H}. The pixel height \textit{h} of the line connecting key points 5 and 17 can be obtained in real-time, and with the known camera focal length \textit{f}, the depth can be computed using Eq.\ref{equ4}. This enables interaction based on the calculated depth \textit{D}. For depth cameras, to maintain consistency with the anchor size used in monocular depth estimation (the line connecting key points 5 and 17) while simplifying the computation, we measure the depth of the virtual key point 21 as a proxy for the overall hand-to-camera distance.

\subsection{Exoskeleton actions FSM transition strategy}\label{subsec2}

With gesture recognition and distance-trigger algorithms in place, the next step is to design a FSM strategy for controlling exoskeleton movements. FSM is a mathematical model used to describe a finite set of states and the process by which they transition in response to events. It is widely applied in fields such as natural sciences, computer science, automation, and robotics for behavior analysis and modeling\cite{bib48}. FSM has three core attributes: first, the number of system states is finite; second, the system can only be in one state at any given moment; and third, under specific conditions, the system can transition from one state to another. This section leverages FSM to construct the action logic for the exoskeleton, aiming to achieve two primary objectives: (1) simplify the action control process, enabling continuous state transitions triggered by single actions, and (2) standardize action execution to prevent inappropriate movements caused by misoperation or incorrect gesture recognition.

\begin{figure}[h!]
    \centering
    \includegraphics[width=0.47\textwidth]{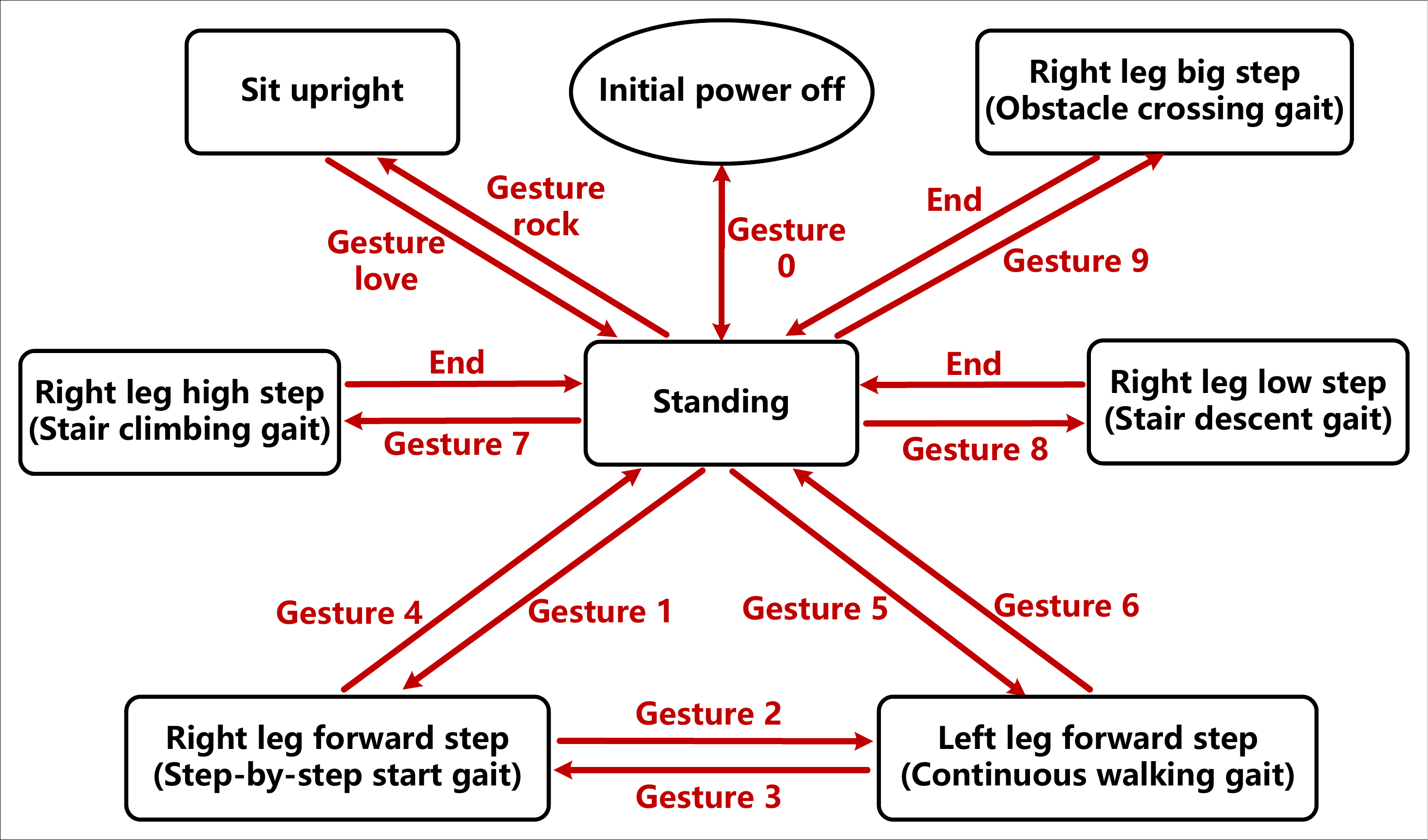}  
    \caption{FSM model of gait switching controlled by gesture in RLEEX}
    \label{fig3}
\end{figure}

The FSM model designed in this section is illustrated in Fig.\ref{fig3}. The system comprises eight states: the initial unpowered state, standing state, sitting state, right leg forward state, left leg forward state, right leg high step state, right leg low step state, and right leg obstacle-crossing state. These states correspond to 12 specific action transitions: transition from the initial unpowered state to the standing state (Gesture 0), transition from the sitting state to the standing state (Gesture rock), transition from the standing state to the sitting state (Gesture love), transition from the standing state to the right leg forward state (Gesture 1), transition from the right leg forward state to the left leg forward state (Gesture 2), transition from the left leg forward state to the right leg forward state (Gesture 3), transition from the right leg forward state to the standing state (Gesture 4), transition from the standing state to continuous walking, starting with the left leg forward (Gesture 5), transition from continuous walking back to the standing state (Gesture 6), transition from the standing state to an ascending stair motion process, starting with the right leg high step, and then returning to the standing state (Gesture 7), transition from the standing state to a descending stair motion process, starting with the right leg low step, and then returning to the standing state (Gesture 8), transition from the standing state to an obstacle-crossing motion process, starting with the right leg, and then returning to the standing state (Gesture 9). Each action is triggered by a gesture, which drives the hip and knee joints to execute corresponding trajectory motions, enabling precise and intuitive control of the exoskeleton system.

The transition between different task states requires executing the corresponding gait actions based on gesture commands combined with virtual button trigger conditions. Specifically:

(1)The initialization of the exoskeleton, stair ascending, stair descending, and obstacle-crossing tasks are triggered upon receiving user Gestures 0, 7, 8, and 9, respectively. Each gesture triggers the execution of the associated gait trajectory. According to the designed gait trajectories for each task, the exoskeleton returns to the standing state upon completing the respective gait tasks.

(2)Users can initiate the exoskeleton's step-by-step walking task from a standing state by performing Gesture 1 and pressing the virtual button, prompting the exoskeleton to begin the right-leg stepping motion and transition from standing to taking a single step. During the step-by-step walking task, the wearer can use Gestures 2 and 3 to control the exoskeleton for alternate left and right leg walking training. Finally, Gesture 4 triggers the retraction motion, returning the exoskeleton to the standing state. Alternatively, users can directly trigger a continuous walking task starting with the left leg forward by performing Gesture 5. Gesture 6 allows users to stop walking at any point, initiating a retraction motion to return to the standing state.

(3)After the exoskeleton completes initialization and transitions to the standing state, users can trigger the “sit down” action by performing Gesture love. Similarly, the exoskeleton in the sitting position can be returned to the standing state by performing Gesture rock.

To ensure user safety, the exoskeleton temporarily ignores gesture commands while executing a gait task until the action is completed. If the wearer experiences discomfort, an emergency stop button can immediately terminate the exoskeleton's movement. Additionally, a majority voting mechanism is employed to address issues of misoperation caused by fluctuations in gesture recognition, enhancing the robustness of interaction control. The FSM processes gesture commands only after receiving the same gesture input across three consecutive frames. Upon releasing the virtual button, the FSM issues a motion command to the exoskeleton based on the current task state.

\section{Experiment and results}\label{sec4}

\subsection{Experimental platform}\label{subsec2}

The experiments were conducted using the Windows 10 operating system as the training platform, implemented with TensorFlow 2.1.0, Python 3.7, and PyTorch 1.7.0. All training was performed on a desktop computer equipped with an Intel(R) Core(TM) i5-10400F CPU @ 2.90GHz and an NVIDIA GeForce GTX 1080Ti GPU. Testing was carried out on an exoskeleton control platform running on a laptop with an Intel(R) Core(TM) i5-4210U CPU @ 1.70GHz. For visual sensors, the built-in RGB camera of the laptop was used as the monocular camera for the experiments, while the Intel(R) RealSense(TM) D435i depth camera was employed for comparison with the monocular RGB camera. Regarding the exoskeleton, we designed and built a lower-limb rehabilitation exoskeleton featuring Quasi-Direct Drive (QDD) actuated joints to validate the proposed interaction control method.

\begin{figure}[h!]
    \centering
    \includegraphics[width=0.48\textwidth]{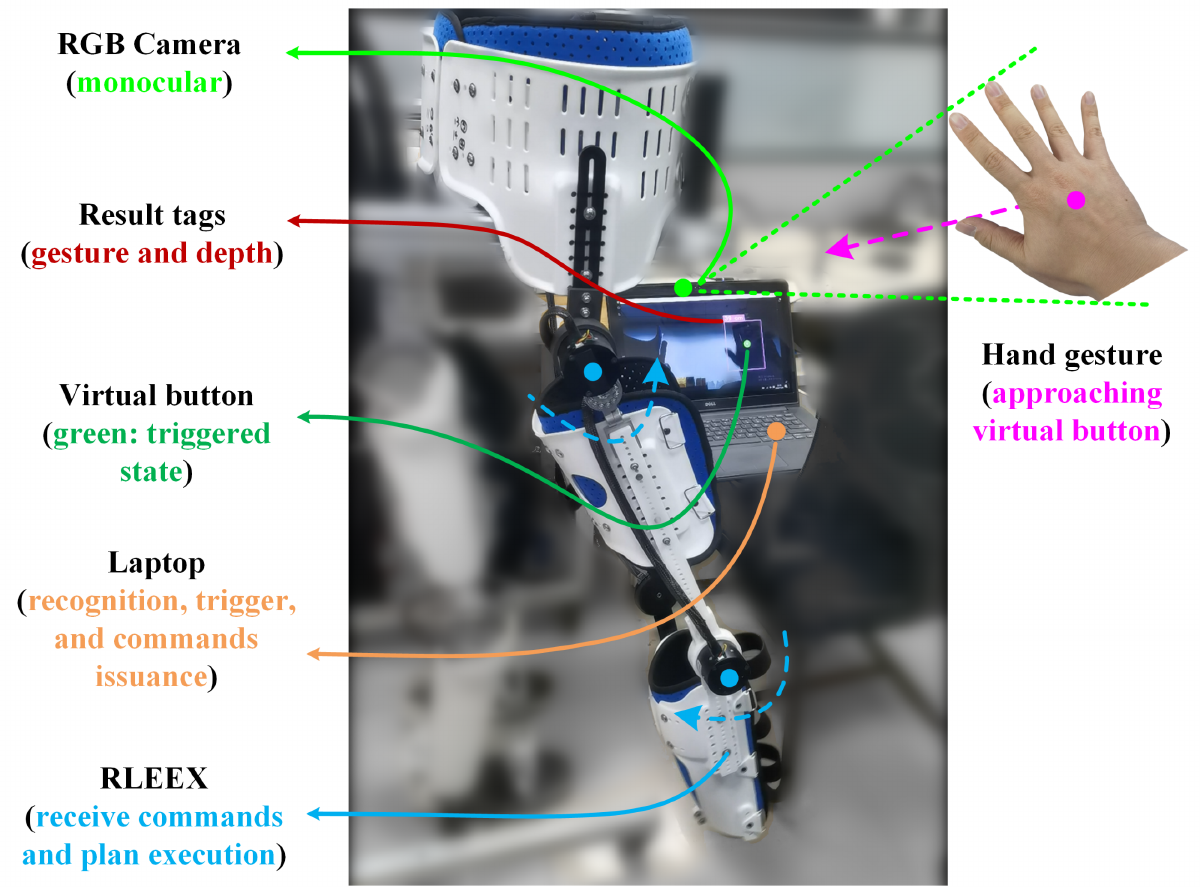}  
    \caption{Physical diagram of the gesture-triggered control RLEEX system}
    \label{fig4}
\end{figure}

Fig.\ref{fig4} illustrates the entire system. In this diagram, the operator controls the RLEEX motion modes using gestures and virtual buttons, addressing the challenges of intent recognition in complex motion modes and the low reliability of visual interaction. The operator's gesture and distance information are captured using the laptop's built-in camera. A supervisory program, written in Python, processes the gesture data, converting it into corresponding gait motion commands for the RLEEX, with the distance information serving as a trigger for interaction. Finally, the control commands are sent to the RLEEX, executing the specific gait actions.

\subsection{Accuracy evaluation of gesture recognition}\label{subsec2}

The accuracy of gesture recognition is directly related to the stability of the entire system. Before evaluating the overall detection performance, the performance of each individual gesture should be assessed. For hand presence, images with annotated regions are defined as positive samples, while images without annotated regions are defined as negative samples. Based on the classifier model, four possible outcomes can be produced: True Positive (TP), True Negative (TN), False Positive (FP), and False Negative (FN). This section evaluates the model's performance using five metrics: Precision (Pre), Recall (Re), Intersection over Union (IoU), F1-Score (F1), and Average Precision (AP). The methods for calculating these metrics are as follows:

\begin{equation}
\text { \textit{Pre} }=\frac{T P}{T P+F P} \times 100 \%
\label{equ5}
\end{equation}

\begin{equation}
R e=\frac{T P}{T P+F N} \times 100 \%
\label{equ6}
\end{equation}

\begin{equation}
I o U=\frac{T P}{T P+F P+F N} \times 100 \%
\label{equ7}
\end{equation}

\begin{equation}
F 1=\frac{2 \cdot \operatorname{\textit{Pre}} \cdot R e}{\text { \textit{Pre} }+R e} \times 100 \%
\label{equ8}
\end{equation}

\begin{equation}
A P=\int_{0}^{1} \operatorname{\textit{Pre}}(\operatorname{\textit{Re}}) \mathrm{d} R e \times 100 \%
\label{equ9}
\end{equation}

It can be observed that Precision is the proportion of actual positive samples among those predicted as positive, while Recall is the proportion of actual positive samples correctly predicted as positive. The Intersection over Union (IoU) is the ratio of the intersection of the predicted set and the true set to their union. The F1-Score is the harmonic mean of Precision and Recall, with a maximum value of 1 and a minimum value of 0. Average Precision (AP) combines both Precision and Recall, providing a comprehensive evaluation of their performance. Furthermore, the overall metric for evaluating the performance of an object detection model is the mean Average Precision (mAP), which integrates the average Precision of the model's detections and the average performance across all classes. The calculation method for mAP is as follows:
\begin{equation}
m A P=\frac{\sum_{i=1}^{k} A P_{i}}{k} \times 100 \%
\label{equ10}
\end{equation}

In the formula, \textit{k} represents the total number of gesture categories, and $A P_{i}$ denotes the accuracy of the \textit{i-th} category, where \textit{i} is the sequence number. This paper uses the mAP at an IoU threshold of 0.5 (where samples with an IoU greater than 0.5 with positive samples are considered True Positives, or TP) as the evaluation metric, referred to as mAP@0.5.

Tab.\ref{Tab3} presents the evaluation results of the model’s predictions for each gesture. It can be observed that Gestures 0 to 6, along with Gestures love and rock, achieved an AP of over 95\%, indicating that the KGDFA model has excellent discrimination for these gestures, with Gesture 5 demonstrating the highest level of distinction. The AP for Gestures 7 and 9, however, is slightly lower. This is due to these gestures having different spatial angles in 3D poses, which significantly affect the accuracy of joint skeleton angles and keypoints distance calculations when projected onto the 2D plane, leading to a decline in recognition precision and other related metrics. Although the evaluation results for these gestures are relatively lower compared to others, their AP remains above 85\%, which is sufficient to distinguish and recognize gestures for our gesture recognition and trigger control tasks.

\begin{table*}[h!]
\centering
\small  
\caption{Evaluation results of KGDFA model for different gesture predictions(Bold indicates the best result)}  
  \label{Tab3}
  \renewcommand{\arraystretch}{1.6}  
  \begin{tabularx}{\textwidth}{>{\centering\arraybackslash}m{2.2cm}|>{\centering\arraybackslash}m{2.2cm}|>{\centering\arraybackslash}m{2.2cm}|>{\centering\arraybackslash}m{2.2cm}|>{\centering\arraybackslash}m{2.2cm}|>{\centering\arraybackslash}m{2.2cm}}  
  \hline
  \textbf{Gesture Name} & \textbf{Pre(\%)} & \textbf{Re(\%)} & \textbf{IoU(\%)} & \textbf{F1(\%)} &  \textbf{AP(\%)} \\
  \hline
  Gesture 0 & 97.92 & 93.31 & 96.88 & 93.13 & 96.02 \\ 
  \hline
  Gesture 1 & 98.43 & 87.73 & 98.10 & 87.75 & 96.49 \\
  \hline
  Gesture 2 & 98.99 & 93.74 & 93.86 & 93.63 & 98.89 \\
  \hline
  Gesture 3 & 98.94 & 94.43 & 97.97 & 94.01 & 98.12 \\
  \hline
  Gesture 4 & 96.71 & 83.36 & 95.68 & 89.14 & 96.63 \\
  \hline
  Gesture 5 & \textbf{99.08} & 95.96 & 96.93 & 95.16 & \textbf{99.01} \\
  \hline
  Gesture 6 & 98.97 & 93.26 & \textbf{98.22} & 93.21 & 98.91 \\
  \hline
  Gesture 7 & 89.01 & 76.98 & 88.09 & 76.81 & 88.33 \\
  \hline
  Gesture 8 & 93.71 & 86.11 & 95.82 & 90.31 & 95.58 \\
  \hline
  Gesture 9 & 86.78 & 71.22 & 85.51 & 71.27 & 85.07 \\
  \hline
  Gesture love & 98.86 & 94.72 & 96.05 & 94.16 & 96.87 \\
  \hline
  Gesture rock & 99.03 & \textbf{96.09} & 97.95 & \textbf{95.84} & 98.02 \\
  \hline
\end{tabularx}
\end{table*}

\begin{table*}[h!]
\centering
\small  
\caption{Comparison of different models for gesture recognition(Bold indicates the best result)}  
  \label{Tab4}
  \renewcommand{\arraystretch}{1.6}  
  \begin{tabularx}{\textwidth}{>{\centering\arraybackslash}m{4cm}|>{\centering\arraybackslash}m{2.5cm}|>{\centering\arraybackslash}m{2.5cm}|>{\centering\arraybackslash}m{2.5cm}|>{\centering\arraybackslash}m{2.5cm}}  
  \hline
  \textbf{Network Model} & \textbf{Params(M)} & \textbf{mAP0.5(\%)} & \textbf{US(s)}  &  \textbf{FPS} \\
  \hline
  Faster-RCNN & 137.06 & 90.21 & 0.92 & 1 \\ 
  \hline
  YOLOv3 & 61.53 & 83.03 & 0.25 & 4 \\
  \hline
  SSD & \textbf{26.29} & 84.26 & 0.12 & 8 \\
  \hline
  Mobilenet-yolov4-lite & 41.01 & 85.25 & 0.06 & 17 \\
  \hline
  YOLOv5m & 35.7 & 87.77 & 0.17 & 6  \\
  \hline
  \textbf{KGDFA} & 43.12 & \textbf{95.67} & \textbf{0.03} & \textbf{34}  \\
  \hline
\end{tabularx}
\end{table*}

\begin{figure}[h!]
    \centering
    \includegraphics[width=0.47\textwidth]{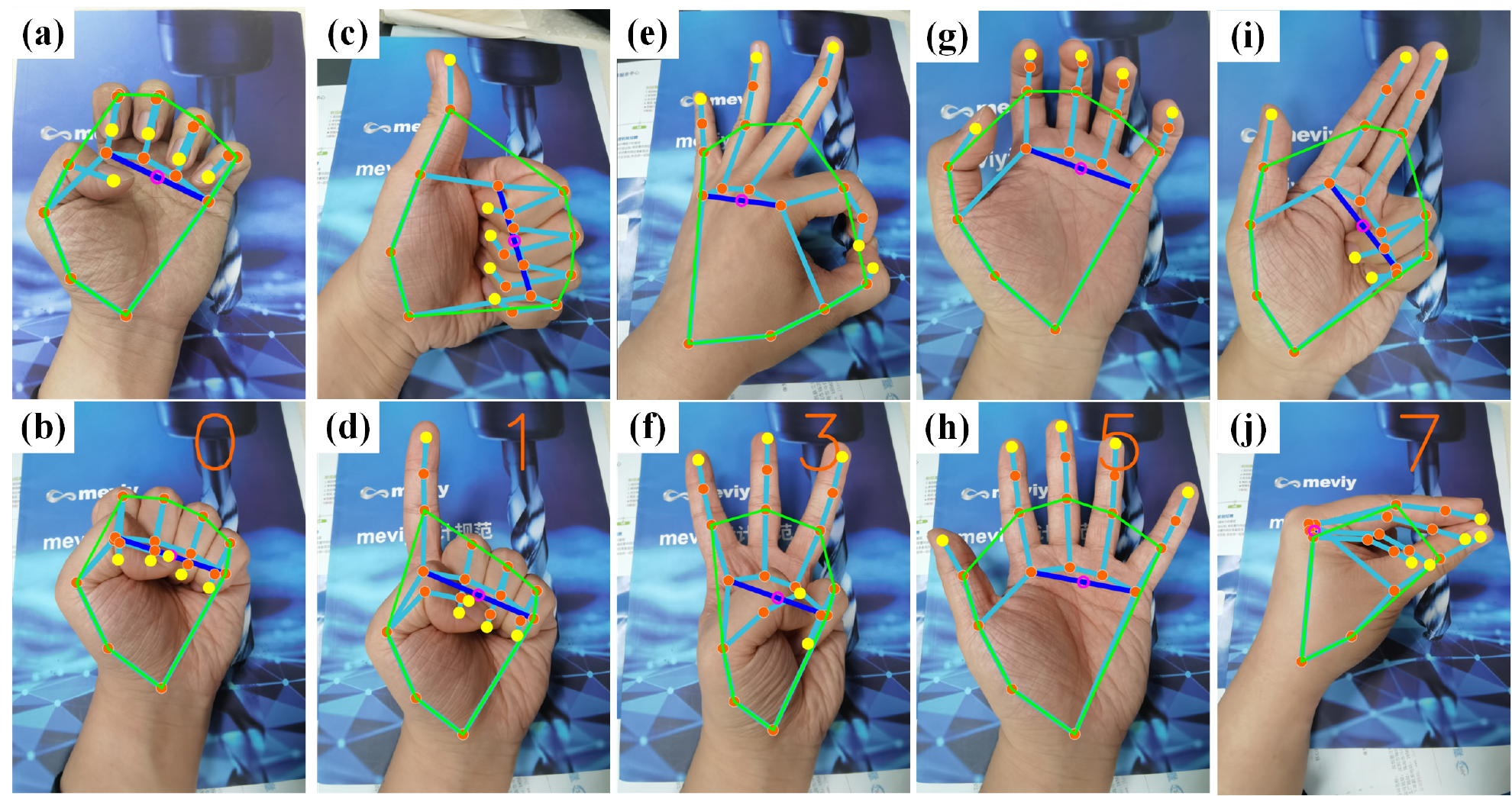}  
    \caption{Confusing gestures:(a) tiger claw hand gesture;(b) defined Gesture 0;(c) good gesture;
(d) defined Gesture 1;(e) ok gesture;(f) defined Gesture 3;(g) naturally open hand gesture;
(h) defined Gesture 5;(i) gun gesture;(j) defined Gesture 7}
    \label{fig5}
\end{figure}

To further validate the superiority of the proposed KGDFA model in recognition tasks, several other algorithm models were selected for comparison with the current model. The comparison models include Faster-RCNN, YOLOv3, SSD, Mobilenet-YOLOv4-lite, and YOLOv5m. We embedded the backbone neural network of the proposed KGDFA model with the comparison models, introducing different features, while maintaining consistent hyperparameter settings during the training phase. In this section:(1)The model complexity is evaluated using the number of parameters (Params). Generally, the smaller the number of parameters, the less storage space the model occupies, making deployment and operation easier.(2)The time taken to process gesture information for a single image (single frame) is used, and the prediction results are assessed by the unit speed (US) metric. A smaller US indicates better model performance.(3)Frames per second (FPS) is used as a real-time performance evaluation metric. A higher FPS indicates lower latency and better real-time performance.

Tab.\ref{Tab4} lists the relevant parameters for various network models. As shown in the table, compared to other algorithms, the KGDFA model proposed in this paper:(1)Achieves a mAP (mAP@0.5) of 95.67\%, outperforming other models, making it the optimal choice for gesture recognition tasks in exoskeleton interaction scenarios.(2)Although the KGDFA model does not have the smallest number of parameters, the time required to process a single frame of gesture information (Unit Speed, US) is only 0.03s, with an FPS of 34, both of which outperform the other models. This indicates that the speed of providing prediction results is sufficient for real-time detection, with low latency, making it suitable for real-time control during the rehabilitation training process using the exoskeleton.

Relying solely on object detection algorithms for gesture recognition can lead to errors, as it lacks constraints on key 2D joint nodes such as the convex surface of the hand, fingertip keypoints, finger flexion degree, pixel distance, and proportions. These missing constraints can easily result in misclassification. As shown in Fig.\ref{fig5}(a), (c), (e), and (g), the presented gestures, including the “tiger claw hand” gesture, “good” gesture, “ok” gesture, and “naturally open hand” gesture, are very similar to the defined Gestures 0, 1, 3, and 5, which may lead to confusion and misclassification. Among these, the “good” and “ok” gestures are commonly used in daily communication and should be avoided to ensure they do not conflict with the exoskeleton gesture control. The constraint fusion algorithm we designed effectively distinguishes between similar gestures by incorporating additional constraints. Undefined gestures will not be recognized. The comparison in Fig.\ref{fig5}(a) and (b) demonstrates how the constraint fusion algorithm successfully differentiates between the “tiger claw hand” gesture and defined Gesture 0 by using the two-layer convex surface recognition constraint. The comparison in Fig.\ref{fig5}(c) and (d), as well as (f) and (h), demonstrates how the constraint fusion algorithm successfully distinguishes the commonly used “good” gesture from the defined Gesture 1, and the “ok” gesture from the defined Gesture 3, by incorporating 2D keypoint constraints on the fingertip positions of all five fingers. The comparison in Fig.\ref{fig5}(g) and (h) demonstrates how the constraint fusion algorithm successfully distinguishes the “naturally open hand” gesture from the defined Gesture 5 by applying constraints based on the degree of finger flexion. The comparison in Fig.\ref{fig5}(i) and (j) demonstrates how the constraint fusion algorithm successfully distinguishes the “gun” gesture from the defined Gesture 7 by applying pixel distance ratio constraints. These results indicate that gestures that are prone to confusion under constraints are accurately recognized, thus preventing misclassification.

Therefore, incorporating 2D joint node constraints such as two-layer convex surface recognition, fingertip keypoints of all five fingers, finger flexion degree, pixel distance, and proportions into the algorithm can significantly improve gesture recognition accuracy. Integrating these steps into the algorithm enhances the robustness of the gesture recognition software, thereby reducing the misclassification of undefined gestures. The correctly recognized gesture data is then converted into control information for the corresponding exoskeleton robot actions. Furthermore, by adopting the two-stage detection approach from MediaPipe (first detecting the palm, then tracking the finger keypoints, with palm detection only running when necessary), real-time control of the exoskeleton's movements during rehabilitation training is effectively achieved.

\subsection{Augmented reality triggered evaluation validation}\label{subsec2}

When using a monocular RGB camera for depth estimation, we conducted comparative validation with a depth camera. A depth camera can calculate the specific depth of a keypoint pixel, which is not feasible with a monocular camera. In this study, the depth camera was used to measure the depth of a virtual point (Point 21) as a proxy for the distance between the hand and the camera, instead of relying on multiple keypoints to determine hand depth. This approach was chosen for three main reasons: (1) Calculating the overall hand depth using the geometric center of all keypoints increases the likelihood of errors, as the geometric center may not lie within the hand itself. (2) Processing multiple keypoints imposes significant computational demands on hardware, reducing processing speed and real-time performance. (3) This method aligns with the monocular depth estimation approach, which uses the line segment between keypoints 5 and 17 as an anchor for scaling. This consistency facilitates a more direct and reliable comparison between the two systems.

\begin{table*}[h!]
\centering
\small  
\caption{Comparison of distance (cm) errors obtained by two cameras in hand-triggered interaction scenarios
(Bold indicates the best result)}  
  \label{Tab5}
  \renewcommand{\arraystretch}{1.6}  
  \begin{tabularx}{\textwidth}{>{\centering\arraybackslash}m{1.40cm}|>{\centering\arraybackslash}m{2.36cm}|>{\centering\arraybackslash}m{2.36cm}|>{\centering\arraybackslash}m{2.36cm}|>{\centering\arraybackslash}m{2.36cm}|>{\centering\arraybackslash}m{2.36cm}}  
  \hline
  \textbf{Ranging number} & \textbf{Actual distance} & \textbf{Monocular distance measurement} & \textbf{Monocular error(\%)}  &  \textbf{Binocular distance measurement} & \textbf{Binocular error(\%)}  \\
  \hline
  1 & 10 & 9.3 & -7\% & / & /  \\ 
  \hline
  2 & 15 & 14.4 & -4\% & / & /  \\
  \hline
  3 & 20 & 19.4 & -3\% & / & /  \\
  \hline
  4 & 25 & 24.3 & -2.8\% & 25.4 & 1.6\%  \\
  \hline
  5 & 30 & 29.5 & -1.6\% & 30.1 & 0.3\%  \\
  \hline
  6 & 35 & 34.5 & -1.4\% & 34.9 & -0.3\%  \\
  \hline
  7 & 40 & 39.1 & -2.25\% & 40.0 & \textbf{0}  \\
  \hline
  8 & 45 & 45.3 & \textbf{0.6\%} & 45.0 & \textbf{0}  \\
  \hline
  9 & 50 & 51.6 & 3.2\% & 50.0 & \textbf{0} \\
  \hline
  10 & 55 & 57.5 & 4.5\% & 54.9 & -0.2\%  \\
  \hline
  11 & 60 & 62.9 & 4.8\% & 59.8 & -0.3\%  \\
  \hline
  12 & 65 & 68.3 & 5.1\% & 64.7 & -0.4\%  \\
  \hline
  13 & 70 & 78.7 & 12.4\% & 69.6 & -0.6\%  \\ 
  \hline
\end{tabularx}
\end{table*}

\begin{table*}[h!]
\centering
\small  
\caption{Comparison of real-time performance and reliability test of two cameras in hand-triggered interaction scenario(Bold indicates the best result)}  
  \label{Tab6}
  \renewcommand{\arraystretch}{1.6}  
  \begin{tabularx}{\textwidth}{>{\centering\arraybackslash}m{1.40cm}|>{\centering\arraybackslash}m{2.36cm}|>{\centering\arraybackslash}m{2.36cm}|>{\centering\arraybackslash}m{2.36cm}|>{\centering\arraybackslash}m{2.36cm}|>{\centering\arraybackslash}m{2.36cm}}  
  \hline
  \textbf{Ranging number} & \textbf{Actual distance} & \textbf{Monocular distance measurement} & \textbf{Monocular error(\%)}  &  \textbf{Binocular distance measurement} & \textbf{Binocular error(\%)}  \\
  \hline
  1 & 10 & 32 & 100\% & \textbf{19} & 100\%  \\ 
  \hline
  2 & 15 & 31 & 100\% & 15 & 100\%  \\
  \hline
  3 & 20 & \textbf{38} & 80\% & 12 & 100\%  \\
  \hline
  4 & 25 & 32 & \textbf{0} & 11 & 90\%  \\
  \hline
  5 & 30 & 33 & \textbf{0} & 10 & 30\%  \\
  \hline
  6 & 35 & 31 & \textbf{0} & 17 & 20\%  \\
  \hline
  7 & 40 & 33 & \textbf{0} & 16 & 20\%  \\
  \hline
  8 & 45 & 37 & \textbf{0} & 11 & 10\%  \\
  \hline
  9 & 50 & 33 & \textbf{0} & 18 & 10\% \\
  \hline
  10 & 55 & 34 & \textbf{0} & 11 & 5\%  \\
  \hline
  11 & 60 & 32 & \textbf{0} & 9 & 5\%  \\
  \hline
  12 & 65 & 35 & \textbf{0} & 11 & \textbf{0}  \\
  \hline
  13 & 70 & 34 & \textbf{0} & 10 & \textbf{0}  \\ 
  \hline
\end{tabularx}
\end{table*}

For our exoskeleton platform, the camera is installed approximately 50 cm away from the body, 25 cm from the same-side hand, and 60 cm from the opposite-side hand. To prevent accidental activation and ensure safety, the virtual button activation distance must be less than the distance to the nearest hand, which is 25 cm. Considering these factors, we set the activation distance threshold to 20 cm. Consequently, the interaction range for hand-triggered interactions on our platform is 20–60 cm. To evaluate the performance of the two cameras in hand-triggered interaction scenarios, we conducted distance measurement tests from a distance slightly larger than the platform's interaction range, increasing in 5 cm increments from near to far. The actual distances measured using a ruler were used as the reference standard. The errors in distance measurements obtained by the two cameras were compared to provide a comprehensive performance assessment. The measurement data and comparison results are shown in Tab.\ref{Tab5}. The results indicate that the monocular camera is significantly less accurate than the depth camera. The depth camera exhibits the largest error at its minimum range, but the error does not exceed 2\%, while errors at other distances are below 1\%. In contrast, the monocular camera generally has errors of 2\% or higher, with a maximum error of 12.4\%. Both cameras perform best at distances of 40–50 cm. At closer distances, accuracy decreases as the distance decreases, and at farther distances, accuracy decreases as the distance increases, indicating a similar trend in depth accuracy between the two cameras. Although the depth camera has higher accuracy, it cannot cover the interaction range required by our platform. Moreover, within the interaction range, the monocular camera's error remains within 5\%, which is acceptable. Therefore, using a monocular camera as the interaction sensor sufficiently meets the requirements for augmented reality-triggered interactions on our platform.

For the interactive screen output at a resolution of 1280×720, we used FPS (frames per second) as the metric to evaluate real-time performance. To assess reliability, we adopted the trigger probability—defined as the ratio of successful triggers to the total number of 20 hand posture changes at a specific distance. The results are presented in Tab.\ref{Tab6}.

The comparison results indicate that the FPS of the depth camera stabilizes between 10 and 12. This limitation arises from the computational burden of obtaining pixel depth, which reduces hardware processing speed and compromises real-time performance. The highest FPS for the depth camera is 19, observed when the hand occupies most of the frame, as the reduced depth computation workload improves processing efficiency. In contrast, the monocular camera consistently achieves an FPS of 31 or higher, significantly surpassing the maximum FPS of the depth camera and demonstrating excellent real-time performance.

\begin{figure}[h!]
    \centering
    \includegraphics[width=0.47\textwidth]{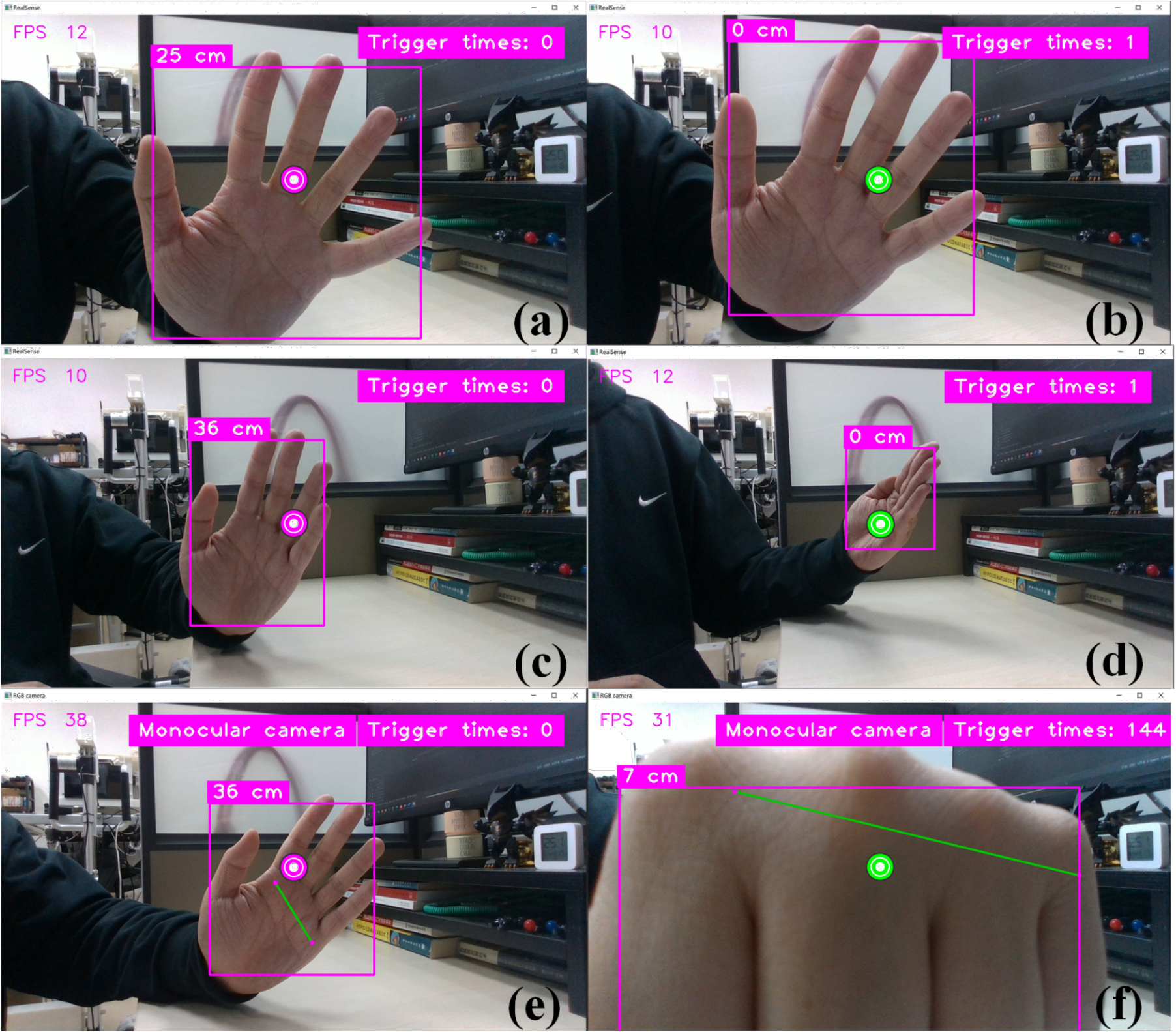}  
    \caption{Special cases of real-time and reliability of two cameras in trigger interaction scenarios(Center virtual button:When it turns green, it means the button is pressed, and when it turns red, it means the button is popped up.):(a) Realsense can measure a minimum depth of 25cm,FPS 12;(b) Realsense returns a measurement value of 0 when it is just less than 25cm,FPS 10;(c) Realsense normally obtains the depth of the palm at 36cm,FPS 10;(d) Realsense incorrectly returns a value of 0 within the measurement range to trigger a virtual key,FPS 12;(e) RGB monocular normal acquisition of palm depth 36cm,FPS 38;(f) RGB monocular can measure a minimum depth of 7cm,FPS 31}
    \label{fig6}
\end{figure}

Additionally, we compared specific cases of the two cameras in the hand-triggered interaction scenario, as shown in Fig.\ref{fig6}. The virtual button activation condition is set such that the hand-to-camera distance must be less than a specified threshold (20 cm in this example). From the binocular trigger probabilities in Tab.\ref{Tab6} and Fig.\ref{fig6}(a) and (b), it can be observed that the depth camera cannot measure distances less than 25 cm and instead returns a depth value of 0, which completely activates the virtual button. Moreover, when the hand is at the edge of the video frame, when there is a sudden change in background depth behind the hand, or when the hand posture changes significantly—particularly during flipping or opening/closing motions—the depth camera may also erroneously return a depth value of 0. This occurs because the infrared laser dot array becomes sparse at the edges, and hand-flipping motions may cause the laser dots to miss the hand entirely and fall into gaps, resulting in false activations, as illustrated in Fig.\ref{fig6}(d). These erroneous depth values of 0 from the depth camera can falsely trigger the virtual button, thereby reducing the reliability of non-contact hand-controlled exoskeleton interactions and increasing safety risks. We also observed that the probability of false activations decreases as the distance from the depth camera increases, and no false activations occur beyond 65 cm. This behavior is determined by the characteristics of the infrared laser dot array: at close distances, gaps are more likely to occur, while at farther distances, the target object appears smaller in the frame, and the laser dot measurements become more consistent. However, such distances exceed the interaction range of our platform.

When using a monocular camera, there is a 20\% probability of failing to trigger the virtual button during hand posture changes near the threshold. This is because changes in hand posture alter the anchor size in the image. The anchor size is defined as the size of the hand when it is perpendicular to the camera, representing the maximum observable size at that distance. Consequently, any posture change reduces the anchor size, leading to overestimated depth measurements. This characteristic ensures that hand movements, such as flipping, at depths greater than the threshold will not falsely trigger the virtual button, as reflected in the monocular trigger probabilities in Tab.\ref{Tab6}. This guarantees the reliability of virtual button activation during interactions. Furthermore, a comparison of Fig.\ref{fig6}(b), (d), and (f) reveals that the monocular camera only loses the palm anchor box when the palm depth is less than 7 cm (i.e., when the hand completely blocks the frame). This demonstrates that the monocular camera has a better dynamic measurement range than the depth camera. Additionally, its depth measurements remain continuously variable without returning erroneous zero values, eliminating the risk of false activations. This enhances the safety and reliability of non-contact hand-controlled exoskeleton interactions.

Of course, monocular camera-based distance measurement also has its challenges, with the most significant being its relatively large errors. This issue is particularly pronounced when the hand is angled relative to the camera, causing the recognized hand area to appear smaller. The reference anchor size also decreases due to the change in perspective, leading to greater measurement inaccuracies. However, in such cases, the monocular camera tends to mistakenly interpret the hand as being farther away, consistent with the pinhole imaging principle of "near objects appear larger, far objects appear smaller." Therefore, these errors do not affect the functionality of approaching and pressing the virtual button.

\subsection{HRI experimental evaluation for gesture control of RLEEX}\label{subsec2}

To calculate the total time from performing a gesture to the initiation of the exoskeleton’s action and to determine the actual success recognition rate, we selected three participants (A, B, and C) to perform 50 independent experiments for each control gesture. All participants were informed of the correct gesture features and the corresponding actions of the rehabilitation exoskeleton robot. Participant A was involved in the network design and training described in this study. To ensure an impartial third-party evaluation, participants B and C were invited and had no prior involvement in any related experiments.

In the practical process of gesture-based interaction control, the exoskeleton robot’s operation commands exhibit slight delays during execution. These delays arise from communication latencies in transmitting recognized trigger commands from the upper-level controller to the lower-level motor actions, including delays in both WiFi and CAN communication. To accurately measure the time required for gesture-based interaction control, we used recorded videos to ensure that the operator's gestures, the interaction recognition screen on the computer, and the entire exoskeleton system were all effectively captured within the camera's field of view. In the video analysis, the first frame marked the recognition of the operator’s gesture, the second frame identified the successful activation of the model (with the virtual button turning from red to green), and the third frame indicated the initiation of the corresponding gait movement by the exoskeleton robot. By conducting multiple measurements and calculating the average difference between the third and first frames, we determined the Average Response Time (ART), representing the mean time required for a gesture to control the exoskeleton robot’s motion.

The primary focus of the evaluation experiment was to test the number of successfully recognized gestures and the ART of the exoskeleton robot. As shown in Tab.\ref{Tab7}, the lowest success recognition rate among the three participants was 80.67\%, observed for Gesture 9. This recognition rate, combined with the augmented reality trigger conditions, meets the usage requirements. The highest success recognition rate reached 97.33\%, demonstrating the robustness of the method and its ability to accurately recognize gestures from different individuals.

The ART for the entire process, from initiating a gesture and triggering the virtual button to the robot’s response, ranged from 0.99 seconds at its slowest to 0.44 seconds at its fastest. These results strongly validate the effectiveness of the proposed KGDFA model in enabling real-time HRI to control exoskeleton gait actions.

\begin{table*}[h!]
\centering
\small  
\caption{Test results of accuracy and real-time performance of gesture control(Bold indicates the best result)}  
  \label{Tab7}
  \renewcommand{\arraystretch}{1.6}  
  \begin{tabularx}{\textwidth}{>{\centering\arraybackslash}m{2cm}|>{\centering\arraybackslash}m{3cm}|>{\centering\arraybackslash}m{3cm}|>{\centering\arraybackslash}m{3cm}|>{\centering\arraybackslash}m{3cm}}  
  \hline
  \textbf{Gesture Name} & \textbf{Number of tests for the same gesture} & \textbf{Average number of successful recognitions by three testers} & \textbf{Accuracy(\%)}  &  \textbf{ART(s)}  \\
  \hline
  Gesture 0 & 50 & 44.67(42/48/44) & 89.33 & 0.49  \\ 
  \hline
  Gesture 1 & 50 & 46.33(48/47/44) & 92.67 & 0.50  \\
  \hline
  Gesture 2 & 50 & 46.67(46/45/\textbf{49}) & 93.33 & 0.53  \\
  \hline
  Gesture 3 & 50 & 46.67(47/45/48) & 93.33 & 0.68  \\
  \hline
  Gesture 4 & 50 & 47(48/46/47) & 94.00 & 0.51  \\
  \hline
  Gesture 5 & 50 & \textbf{48.67}(\textbf{49}/48/\textbf{49}) & \textbf{97.33} & \textbf{0.44}  \\
  \hline
  Gesture 6 & 50 & 47.67(48/46/\textbf{49}) & 95.33 & 0.46 \\
  \hline
  Gesture 7 & 50 & 41.33(44/41/39) & 82.67 & 0.89  \\
  \hline
  Gesture 8 & 50 & 43.67(46/42/43) & 87.33 & 0.78  \\
  \hline
  Gesture 9 & 50 & 40.33(40/38/43) & 80.67 & 0.99  \\
  \hline
  Gesture love & 50 & 47(46/\textbf{49}/46) & 94.00 & 0.66  \\
  \hline
  Gesture rock & 50 & 48.33(48/\textbf{49}/48) & 96.67 & 0.45  \\
  \hline
\end{tabularx}
\end{table*}

\subsection{Results analysis}\label{subsec2}
Real-time gesture control requires both accurate gesture recognition and synchronized control of the exoskeleton. Further calculations based on the results in Tab.\ref{Tab7} show that the average success recognition rate for the three participants was 94.11\%, and the total average response time for the system to complete exoskeleton hand gesture control was approximately 0.615 seconds. These findings indicate that the proposed method not only achieves a high success rate and excellent real-time performance but also emphasizes its effectiveness in enabling non-experts to control the robot in real time. The experimental results demonstrate the potential of combining gesture recognition technology, depth estimation-based augmented reality techniques, and rehabilitation exoskeleton control technologies.

The proposed KGDFA model introduces a novel approach to address the challenges encountered in HRI for rehabilitation exoskeletons, including the complex and diverse recognition of human motion intentions and ensuring a reliable, sensitive, and comfortable user experience. Additionally, it enhances the interactive aspect of the rehabilitation training process, making it more engaging for users. This increased interactivity helps to improve patients' willingness and enthusiasm to participate in their rehabilitation training, encouraging active involvement in the process.

\section{Summary and future research}\label{sec5}

Gesture recognition forms the foundation of gesture-based control and is crucial in HRI environments. Once accurate gesture recognition is achieved, the gesture information can be converted into the corresponding gait movement information for the RLEXX exoskeleton. Distance data serves as the trigger to control the activation of the appropriate gait movement in RLEXX. The time required for gesture recognition and triggering the response is relatively short, allowing for real-time interaction.

To address the issues of discomfort, lack of intuitiveness, and unnaturalness associated with contact-based HRI on the RLEEX platform, this paper proposes a non-contact KGDFA real-time interactive control method for RLEEX. This approach not only provides a novel perspective for controlling RLEEX but also offers several advantages over traditional control methods. First, traditional HRI systems typically rely on contact sensors to gather information, which is then fed back to the controller for action. Prolonged interaction with these sensors can lead to physical discomfort for the user. In contrast, the non-contact HRI method proposed in this paper allows users to control the robot remotely, providing a more intuitive, natural, and comfortable experience. Second, traditional visual gesture recognition systems often lack specific trigger conditions, making them prone to false activations, which can reduce the reliability and safety of the interactive control system. The virtual button functionality designed in this study introduces determinism into the interaction process, ensuring the safety of the interaction. Finally, the KGDFA control method proposed here is based on a monocular RGB camera, eliminating the need for a depth camera. This not only reduces costs but also ensures broader applicability. Experimental results further validate the effectiveness and reliability of the RLEEX gesture-based interactive control system.

Nevertheless, the method proposed in this study has some limitations: (1) The camera used for gesture recognition has strict requirements regarding the surrounding environment and lighting conditions. (2) The current KGDFA real-time control system experiences latency issues, resulting in slight stuttering during HRI. This prevents rapid adjustments of gait movements through frequent gesture modifications.

Future research should prioritize addressing the aforementioned shortcomings and seek to develop a non-contact real-time control system with broader environmental applicability, lower latency, and greater reliability. This will enhance the universality, smoothness, and safety of gesture-based control for the RLEEX exoskeleton. To better apply the non-contact HRI technology proposed in this paper to a wider range of environments, we plan to introduce multi-sensor fusion techniques, such as voice recognition, into the gesture interaction control system. This will enhance robustness against environmental changes. Additionally, we aim to further improve the real-time performance of the interaction control by optimizing recognition, triggering, and communication algorithms. In summary, the three-step fusion HRI control algorithm based on non-contact sensors for keypoints detection, gesture classification, and distance triggering holds significant potential for broad application and future development in the RLEEX exoskeleton.

\section*{Acknowledgements}

We wish to express our gratitude to the participants who took part in this article.

\section*{Data Availability}

Datasets for this article are available at MSRA Hand Tracking database (\url{https://www.dropbox.com/s/t91imizfdaf4i5l/cvpr14\_MSRAHandTrackingDB.zip?dl=0}) and MSRA Hand Gesture database (\url{https://www.dropbox.com/s/bmx2w0zbnyghtp7/cvpr15\_MSRAHandGestureDB.zip?dl=0}).

\section*{Author Contributions}

\textbf{Shuang Qiu}: Conceptualization, Methodology, Experiment, Analysis, Writing - original draft. \textbf{Zhongcai Pei}: Discussion, Funding acquisition, Supervision. \textbf{Chen Wang}: Discussion, Deep learning neural network construction and recurrence, Writing - review \& editing. \textbf{Jing Zhang}: Writing - translation \& polishing. \textbf{Zhiyong Tang}: Discussion, Suggestion, Funding acquisition, Supervision

\section*{Declarations}

Conflict of Interests: The authors have no conflicts of interest to declare that are relevant to the content of this article.Ethics Approval: Not applicable.

\bibliographystyle{unsrt}
\bibliography{sn-bibliography.bib}


\end{document}